\documentclass[final]{cvpr}

\usepackage{times}
\usepackage{epsfig}
\usepackage{graphicx}
\usepackage{amsmath}
\usepackage{amssymb}
\usepackage{subcaption}
\usepackage{multirow}
\usepackage{indentfirst}
\usepackage{pifont}
\usepackage{bm}
\usepackage{float}

\usepackage[pagebackref=true,breaklinks=true,colorlinks,bookmarks=false]{hyperref}

\def\cvprPaperID{2471} 
\def\confYear{CVPR 2021}

\begin{document}

\title{Objects are Different: Flexible Monocular 3D Object Detection}
\author{Yunpeng Zhang, Jiwen Lu\thanks{Corresponding author}, Jie Zhou\\
Beijing National Research Center for Information Science and Technology, China\\
Department of Automation, Tsinghua University, China\\
{\tt \small zhang-yp19@mails.tsinghua.edu.cn; \{lujiwen,jzhou\}@tsinghua.edu.cn}
}
\maketitle

\thispagestyle{empty}
\pagestyle{empty}

\begin{abstract}
The precise localization of 3D objects from a single image without depth information is a highly challenging problem. Most existing methods adopt the same approach for all objects regardless of their diverse distributions, leading to limited performance for truncated objects. In this paper, we propose a flexible framework for monocular 3D object detection which explicitly decouples the truncated objects and adaptively combines multiple approaches for object depth estimation. Specifically, we decouple the edge of the feature map for predicting long-tail truncated objects so that the optimization of normal objects is not influenced. Furthermore, we formulate the object depth estimation as an uncertainty-guided ensemble of directly regressed object depth and solved depths from different groups of keypoints. Experiments demonstrate that our method outperforms the state-of-the-art method by relatively 27\% for the moderate level and 30\% for the hard level in the test set of KITTI benchmark while maintaining real-time efficiency. Code will be available at \url{https://github.com/zhangyp15/MonoFlex}.
\end{abstract}

\section{Introduction}
3D object detection is an indispensable premise for machines to perceive the physical environment and has been widely used in autonomous driving and robot navigation.
In this work, we focus on solving the problem with only information from monocular images. Most existing methods for 3D object detection require the LiDAR sensors~\cite{lang2019pointpillars, maturana2015voxnet, qi2018frustum, shi2020pv, shi2019pointrcnn, yan2018second} for precise depth measurements or stereo cameras~\cite{3DOP, li2019stereo, TL-stereo, wang2019pseudo} for stereo depth estimation, which greatly increases the implementation costs of practical systems. Therefore, monocular 3D object detection has been a promising solution and received much attention in the community~\cite{barabanau2019monocular, Brazil_2019_ICCV, mono3d, chen2020cvpr, ding2020learning, ku2019monopsr, liu2019deep, Ma_2019_ICCV, deep3Dbox}.

\begin{figure}[t]
   \centering
   \begin{subfigure}{0.49\linewidth}
   \includegraphics[width=\linewidth]{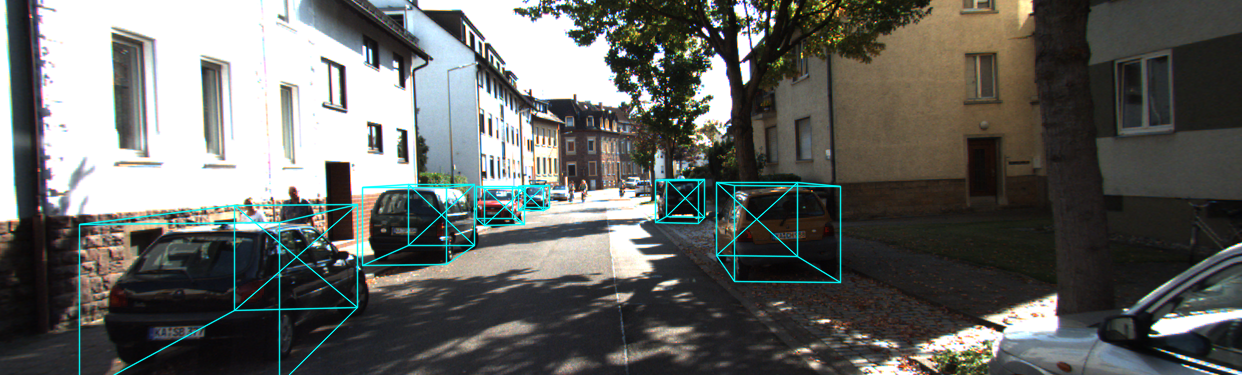}
   \caption{M3D-RPN~\cite{Brazil_2019_ICCV}}
   \end{subfigure}
   \begin{subfigure}{0.49\linewidth}
   \includegraphics[width=\linewidth]{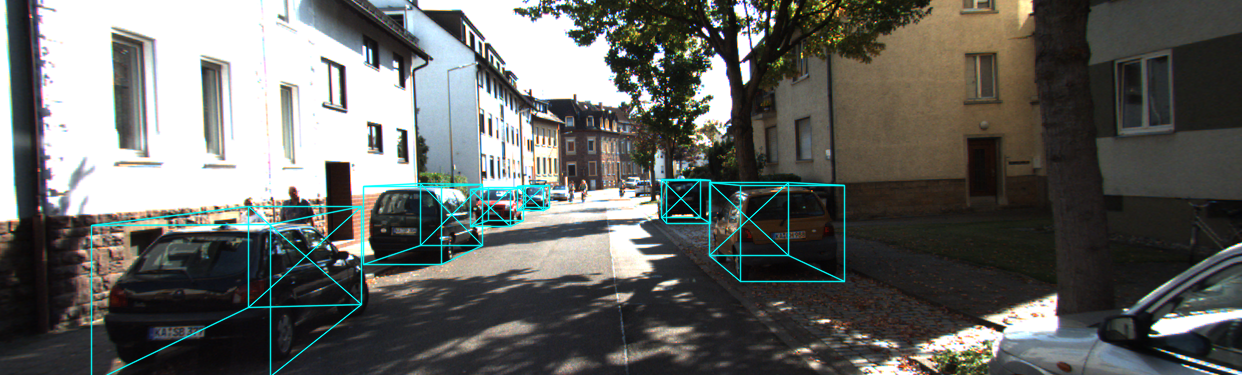}
   \caption{D4LCN~\cite{ding2020learning}}
   \end{subfigure}
   \begin{subfigure}{0.49\linewidth}
   \includegraphics[width=\linewidth]{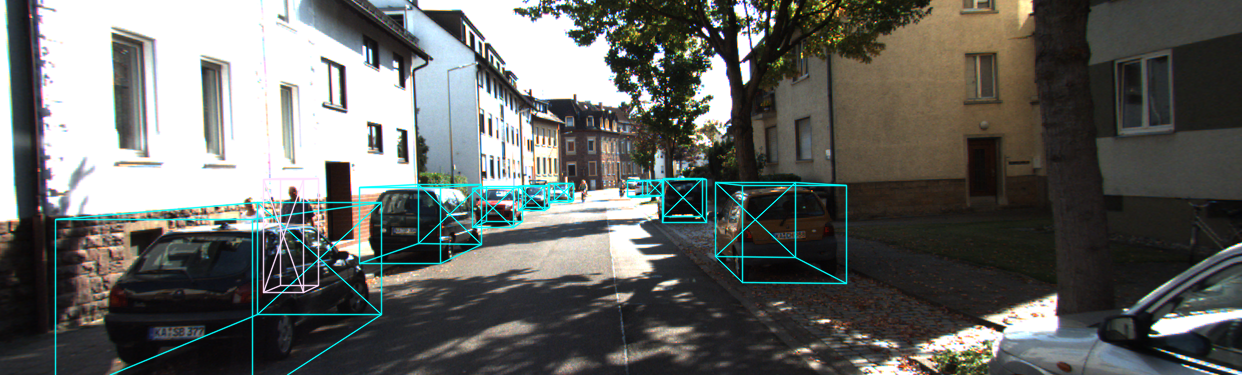}
   \caption{Baseline}
   \end{subfigure}
   \begin{subfigure}{0.49\linewidth}
   \includegraphics[width=\linewidth]{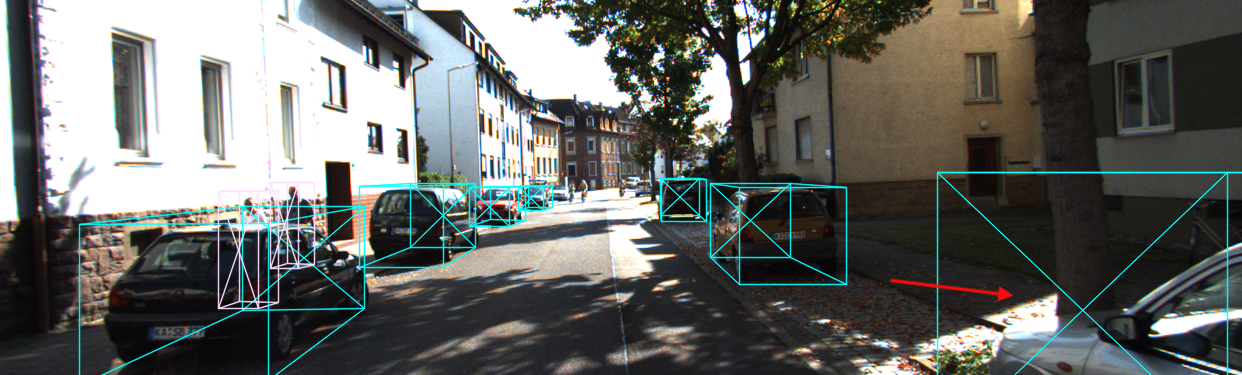}
   \caption{Ours}
   \end{subfigure}
   \caption{Qualitative comparison among prior arts~\cite{Brazil_2019_ICCV, ding2020learning}, our baseline, and the proposed method. The cyan and pink bounding boxes represent detected cars and pedestrians. Our approach can effectively detect the heavily truncated object highlighted by the red arrow.}
   \label{fig:motivation}
\end{figure}

For the challenging localization of 3D objects, most existing methods handle different objects with a unified approach. For example, \cite{chen2020cvpr, li2020rtm3d, liu2020smoke, zhou2019objects} utilize fully convolutional nets to predict objects of diverse distributions with shared kernels. However, we observe that the equal and joint processing of all objects can lead to unsatisfied performance: (1) As shown in Figure~\ref{fig:motivation}, the heavily truncated objects can be hardly detected by state-of-the-art methods~\cite{Brazil_2019_ICCV, ding2020learning} but these objects are important to the safety of autonomous vehicles. (2) We empirically found that these hard samples can increase the learning burden and affect the prediction of general objects. Thus, unified approaches can fail in both finding every object and predicting precise 3D locations. To this end, we propose a flexible detector that considers the difference among objects and estimates their 3D locations in an adaptive way.
Since the estimation of an object's 3D location is usually decomposed into finding the projected 3D center and the object depth~\cite{chen2020cvpr, liu2020smoke, qin2019monogrnet, zhou2019objects}, we also consider the flexibility from these two aspects.

To localize the projected 3D center, we divide objects according to whether their projected centers are ``inside" or ``outside" the image. Then we represent inside objects exactly as the projected centers and outside objects as delicately chosen edge points so that two groups of objects are handled by the inner and edge regions of the feature map respectively.
Considering it is still difficult for convolutional filters to manage spatial-variant predictions, the edge fusion module is further proposed to decouple the feature learning and prediction of outside objects.

To estimate the object depth, we propose to combine different depth estimators with uncertainty estimation~\cite{kendall2017uncertainties, kendall2018multi}.
The estimators include direct regression~\cite{chen2020cvpr, li2020rtm3d, qin2019monogrnet, zhou2019objects} and geometric solutions from keypoints~\cite{barabanau2019monocular, cai2020monocular}.
We observe that computing depth from keypoints is usually an over-determined problem, where simply averaging results from different keypoints~\cite{cai2020monocular} can be sensitive to the truncation and occlusion of keypoints. As a result, we further split keypoints into $M$ groups, each of which is exactly sufficient for solving the depth.
To combine $M$ keypoint-based estimators and the direct regression, we model their uncertainties and formulate the final estimation as an uncertainty-weighted average.
The proposed combination allows the model to flexibly choose more suitable estimators for robust and accurate predictions.


Experimental results on KITTI~\cite{geiger2012we} dataset demonstrate that our method significantly outperforms all existing methods, especially for moderate and hard samples.
The main contributions of this paper can be summarized in two aspects: (1) We reveal the importance to consider the difference among objects for monocular 3D object detection and propose to decouple the prediction of truncated objects; (2) We propose a new formulation for object depth estimation, which utilizes uncertainties to flexibly combine independent estimators. 

\section{Related Work}

\noindent\textbf{Monocular 3D object Detection.}
Considering the difficulty in perceiving 3D environments from 2D images, most existing methods for monocular 3D object detection utilize extra information to simplify the task, which includes pretrained depth estimation modules~\cite{PatchNet, wang2019pseudo, weng2019monocular, xu2018multi}, annotated keypoints~\cite{barabanau2019monocular} and CAD models~\cite{manhardt2019roi}.
Mono3D~\cite{mono3d} first samples candidates based on the ground prior and scores them with semantic/instance segmentation, contextual information, object shape, and location prior.
MonoPSR~\cite{Ku_2019_CVPR} estimates the instance point cloud and enforces the alignment between the object appearance and the projected point cloud for proposal refinement.
Pseudo-LiDAR~\cite{wang2019pseudo} lifts the monocular image into pseudo-LiDAR with estimated depth and then utilizes LiDAR-based detectors.
AM3D~\cite{Ma_2019_ICCV} proposes a multi-modal fusion module to enhance the pseudo-LiDAR with color information. 
PatchNet~\cite{PatchNet} organizes pseudo-LiDAR into the image representation and utilizes powerful 2D CNN to boost the detection performance. 
Though these methods with extra information usually achieve better performance, they require more annotations for training and are usually less generalized.

Other purely monocular methods~\cite{Brazil_2019_ICCV, chen2020cvpr, liu2019deep, liu2020smoke, deep3Dbox, qin2019monogrnet} only utilize a single image for detection. Deep3DBox~\cite{deep3Dbox} presents the \textit{MultiBin} method for orientation estimation and uses geometric constraints of 2D bounding boxes to derive 3D bounding boxes.
FQNet~\cite{liu2019deep} measures the fitting degree between projected 3D proposals and objects so that the best-fitted proposals are picked out.
MonoGRNet~\cite{qin2019monogrnet} directly predicts the depth of objects with sparse supervision and combines early features to refine the location estimation.
M3D-RPN~\cite{Brazil_2019_ICCV} solves the problem with a 3D region proposal network and proposes the depth-aware convolutional layers to enhance extracted features.
MonoPair~\cite{chen2020cvpr} considers the pair-wise relationships between neighboring objects, which are utilized as spatial constraints to optimize the results of detection.
RTM3D~\cite{li2020rtm3d} predicts the projected vertexes of the 3D bounding box and solves 3D properties with nonlinear least squares optimization.
Existing methods mostly neglect the difference among objects or only consider the general scale variance, which can suffer from predicting out-of-distribution objects and lead to downgraded performance. By contrast, our work explicitly decouples the heavily truncated objects with long-tail distribution for efficient learning and estimates the object depth by adaptively combining multiple depth estimators instead of utilizing one single method for all objects.

\begin{figure*}[t]
\centering
\includegraphics[width=0.95\linewidth]{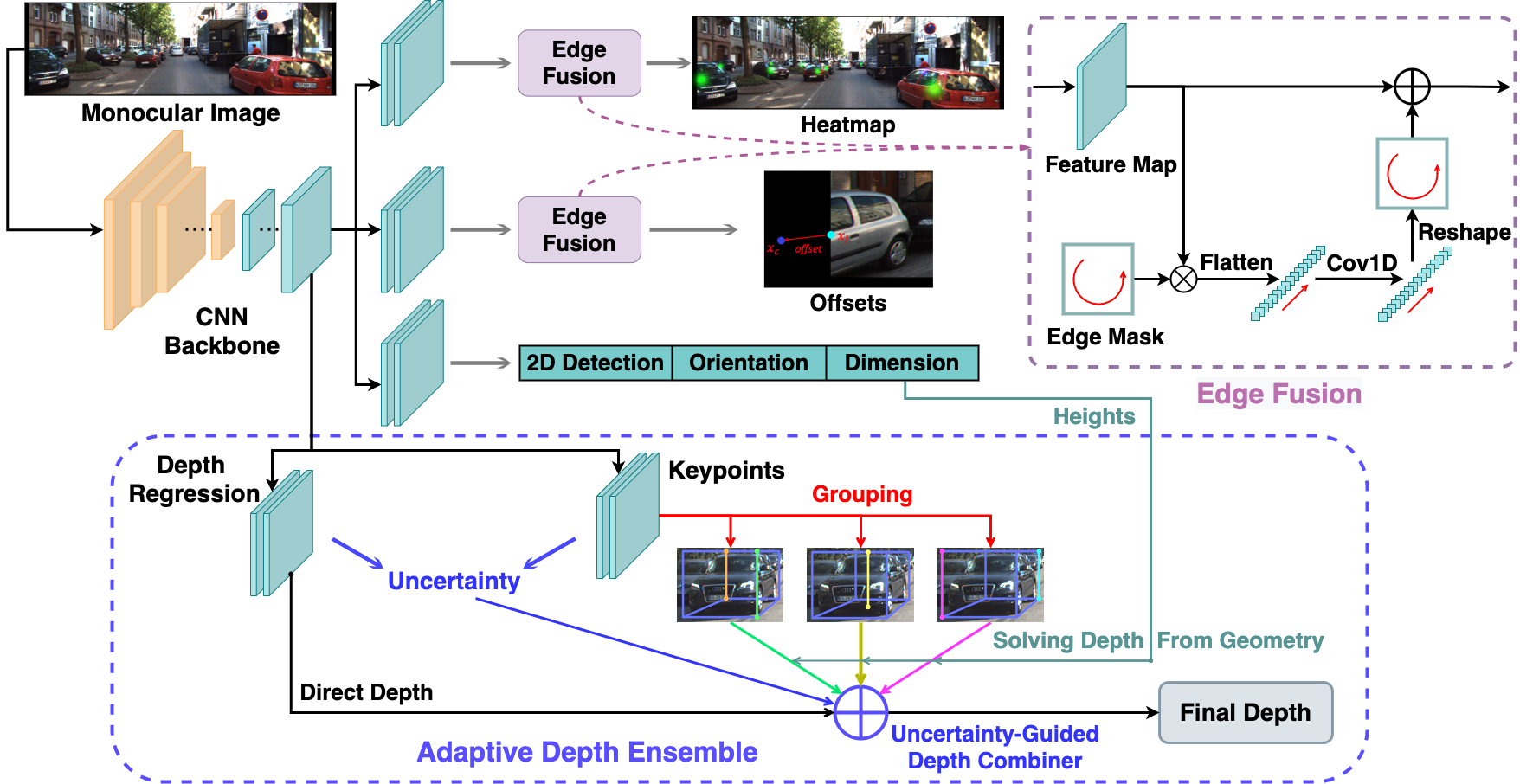}
\caption{Overview of our framework. First, the CNN backbone extracts feature maps from the monocular image as the input for multiple prediction heads. The image-level localization involves the heatmap and offsets, where the edge fusion modules are used to decouple the feature learning and prediction of truncated objects. The adaptive depth ensemble adopts four methods for depth estimation and simultaneously predicts their uncertainties, which are utilized to form an uncertainty-weighted prediction.}
\label{fig:framework}
\end{figure*}

\noindent\textbf{Uncertainty Estimation.}
Two major types of uncertainty are usually studied in Bayesian modeling~\cite{kendall2017uncertainties}. The epistemic uncertainty describes the uncertainty of the model parameters, while the aleatoric uncertainty can capture the noise of observations, whose applications in object detection have been explored in~\cite{chen2020cvpr, choi2019gaussian, he2019bounding}. Gaussian YOLO~\cite{choi2019gaussian} models the uncertainty of predicted 2D boxes to rectify the detection scores.
\cite{he2019bounding} predicts the bounding box as a Gaussian distribution and formulates the regression loss as the KL divergence.
MonoPair~\cite{chen2020cvpr} uses uncertainty to provide weights for the post-optimization between predicted 3D locations and pair-wise constraints.
In this paper, we model the uncertainties of estimated depths from multiple estimators, which are used to quantify their contributions to the final combined prediction.

\noindent\textbf{Ensemble Learning.}
Ensemble learning~\cite{arnaud2019tree, dietterich2002ensemble, jacobs1991adaptive, Mheads, rokach2010ensemble} generates multiple models strategically and combines their predictions for better performance. Traditional ensemble methods include bagging, boosting, stacking, gating network, and so on.
\cite{jacobs1991adaptive} uses a gating network to combine the mixture of experts for classification.
\cite{arnaud2019tree} proposes a tree-structured gate to hierarchically weight different experts for face alignment.
Ensemble learning generally assumes the learners have identical structures but are trained with different samples or initializations, while our multiple depth estimators function in respectively different ways and are also supervised by substantially different loss functions. Therefore, we propose to formulate the combination as an uncertainty-guided average of all predictions.

\section{Approach}
\subsection{Problem Statement}
\begin{figure}[t]
  \centering
  \begin{subfigure}{0.49\linewidth}
  \includegraphics[width=\linewidth]{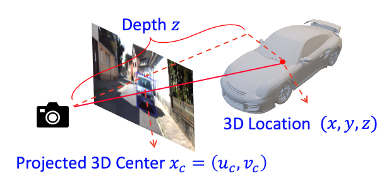}
  \caption{Decomposing 3D locations}
  \end{subfigure}
  \begin{subfigure}{0.49\linewidth}
  \includegraphics[width=\linewidth]{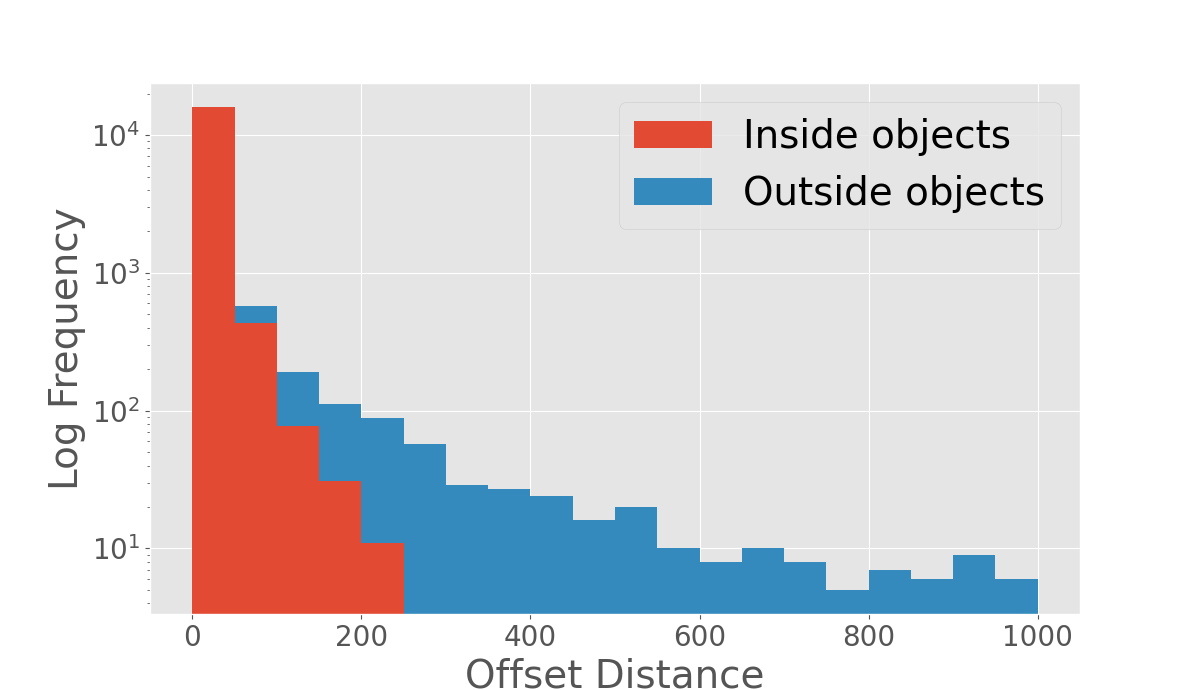}
  \caption{Offset distribution}
  \end{subfigure}
  \caption{(a) The 3D location is converted to the projected center and the object depth. (b) The distribution of the offsets $\bm{\delta_c}$ from 2D centers to projected 3D centers. Inside and outside objects exhibit entirely different distributions.
  }
  \label{fig:offset_distribution}
\end{figure}

The 3D detection of an object involves estimating its 3D location $(x, y, z)$, dimension $(h, w, l)$, and orientation $\theta$. The dimension and orientation can be directly inferred from appearance-based clues, while the 3D location is converted to the projected 3D center $\bm{x_c}=(u_c, v_c)$ and the object depth $z$ as shown in Figure~\ref{fig:offset_distribution}(a) and~\eqref{equ:center_proj}:
\begin{equation}
    x = \frac{(u_c - c_u)z}{f}, \quad y = \frac{(v_c - c_v)z}{f}
    \label{equ:center_proj}
\end{equation}
where $(c_u, c_v)$ is the principle point and $f$ is the focal length. To this end, the whole problem is decomposed into four independent subtasks.

\subsection{Framework Overview}
As shown in Figure~\ref{fig:framework}, our framework is extended from CenterNet~\cite{zhou2019objects}, where objects are identified by their representative points and predicted by peaks of the heatmap. Multiple prediction branches are deployed on the shared backbone to regress objects' properties, including the 2D bounding box, dimension, orientation, keypoints, and depth. The final depth estimation is an uncertainty-guided combination of the regressed depth and the computed depths from estimated keypoints and dimensions.
We present the design of decoupled representative points for normal and truncated objects in Section~\ref{sec:representation} and then introduce the regression of visual properties in Section~\ref{sec:property}. Finally, the adaptive depth ensemble is detailed in Section~\ref{sec:adaptive_depth}.

\subsection{Decoupled Representations of Objects}
\label{sec:representation}
Existing methods~\cite{chen2020cvpr, li2020rtm3d, zhou2019objects} utilize a unified representation $\bm{x_r}$, the center of 2D bounding box $\bm{x_b}$, for every object. In such cases, the offset $\bm{\delta_c} = \bm{x_c} - \bm{x_b}$ is regressed to derive the projected 3D center $\bm{x_c}$. We divide objects into two groups depending on whether their projected 3D centers are \textbf{inside} or \textbf{outside} the image and visualize the corresponding offsets $\bm{\delta_c}$ in Figure~\ref{fig:offset_distribution}(b). Considering the substantially different offsets of two groups, the joint learning of $\bm{\delta_c}$ can suffer from long-tail offsets and we therefore propose to decouple the representations and the offset learning of inside and outside objects.

\noindent\textbf{Inside Objects.}
For objects whose projected 3D centers are inside the image, they are directly identified by $\bm{x_c}$ to avoid regressing the irregular $\bm{\delta_c}$ like~\cite{chen2020cvpr, li2020rtm3d}.
Though we still need to regress the discretization error $\bm{\delta_{in}}$ due to the downsampling ratio $S$ of the backbone CNN as in~\eqref{equ:offset_inside}, it is much smaller than $\bm{\delta_c}$ and easier to regress.
\begin{equation}
   \bm{\delta_{in}} = \frac{\bm{x_c}}{S} - \lfloor \frac{\bm{x_c}}{S} \rfloor
   \label{equ:offset_inside}
\end{equation}

We follow~\cite{zhou2019objects} to generate the ground-truth heatmap for inside objects with circular Gaussian kernels centered at $\bm{x_c}$.

\begin{figure}[t]
\centering
  \begin{subfigure}{0.49\linewidth}
  \includegraphics[width=\linewidth]{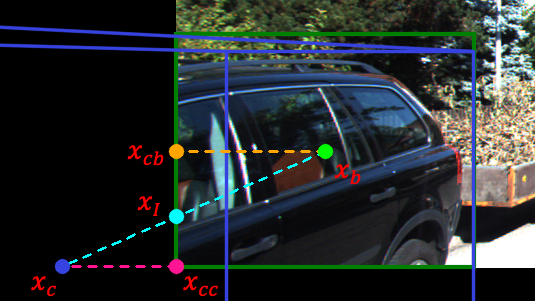}
  \caption{Intersection}
  \end{subfigure}
  \begin{subfigure}{0.49\linewidth}
  \includegraphics[width=\linewidth]{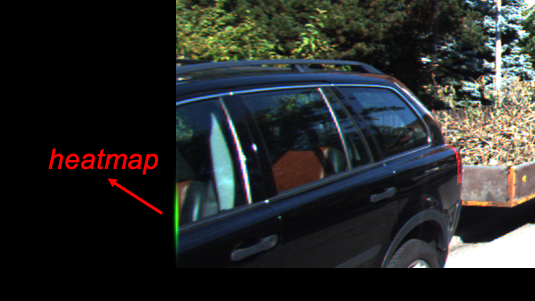}
  \caption{Edge heatmap}
  \end{subfigure}
  \begin{subfigure}{1.0\linewidth}
  \includegraphics[width=\linewidth]{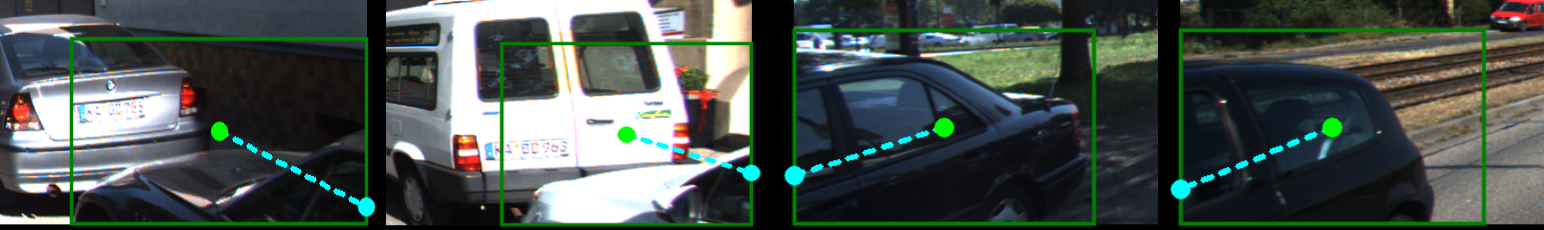}
  \caption{Comparison between $\bm{x_I}$ and $\bm{x_b}$}
  \end{subfigure}
  \caption{Representations of outside objects. (a) The intersection $\bm{x_I}$ between the image edge and the line from $\bm{x_b}$ to $\bm{x_c}$ is used to represent the truncated object. (b) The edge heatmap is generated with 1D Gaussian distribution whose kernel size is proportional to the size of the 2D bounding box. (c) The always-on-edge intersection $\bm{x_I}$ (cyan) is a better representation than the 2D center $\bm{x_b}$ (green) for heavily truncated objects. Best viewed in color.}
  \label{fig:center_fig}
\end{figure}

\noindent\textbf{Outside Objects.}
To decouple the representation of outside objects, we propose to identify them by the intersection $\bm{x_I}$ between the image edge and the line from $\bm{x_b}$ to $\bm{x_c}$, as shown in Figure~\ref{fig:center_fig}(a).
It can be seen that the proposed intersection $\bm{x_I}$ is more physically meaningful than simply clamping $\bm{x_b}$ or $\bm{x_c}$ to the boundary.
The prediction of $\bm{x_I}$ is achieved by the edge heatmap as shown in Figure~\ref{fig:center_fig}(b), which is generated from a one-dimensional Gaussian kernel.
We also compare our $\bm{x_I}$ and the commonly used $\bm{x_b}$ in Figure~\ref{fig:center_fig}(c). Since 2D bounding boxes only capture the inside-image part of objects, the visual locations of $\bm{x_b}$ can be confusing and even on other objects.
By contrast, the intersection $\bm{x_I}$ disentangles the edge area of the heatmap to focus on outside objects and offers a strong boundary prior to simplify the localization. Also, we regress the offsets from $\bm{x_I}$ to the target $\bm{x_c}$ as in~\eqref{equ:offset_outside}:
\begin{equation}
   \bm{\delta_{out}} = \frac{\bm{x_c}}{S} - \lfloor \frac{\bm{x_I}}{S} \rfloor
   \label{equ:offset_outside}
\end{equation}

\noindent\textbf{Edge Fusion.}
Though the representations of inside and outside objects are decoupled in the interior and marginal regions of the output feature, it is still difficult for shared convolutional kernels to handle spatial-variant predictions. Thus, we propose an edge fusion module to further decouple the feature learning and prediction of outside objects.
As shown in the right part of Figure~\ref{fig:framework}, the module first extracts four boundaries of the feature map and concatenates them into an edge feature vector in clockwise order, which is then processed by two 1D convolutional layers to learn unique features for truncated objects. Finally, the processed vector is remapped to the four boundaries and added to the input feature map. When applied to the heatmap prediction, the edge features can specialize in predicting the edge heatmap for outside objects so that the localization of inside objects is not confused. For regressing the offsets, the significant scale difference between $\bm{\delta_{in}}$ and $\bm{\delta_{out}}$ as shown in Figure~\ref{fig:offset_distribution}(b) can be resolved with the edge fusion module.

\noindent\textbf{Loss Functions.}
The penalty-reduced focal loss~\cite{Lin_2017_ICCV} is utilized for heatmap prediction as in~\cite{chen2020cvpr, li2020rtm3d, liu2020smoke}. We adopt L1 loss for regressing $\bm{\delta_{in}}$ and log-scale L1 loss for $\bm{\delta_{out}}$ because it is more robust to extreme outliers. The offset loss is computed as~\eqref{equ:offset_loss}:
\begin{equation}
   L_{off} =
   \left\{
      \begin{aligned}
         & \left| \bm{\delta_{in}} - \bm{\delta_{in}}^* \right| \ \ \text{if inside} \\
         & \log \left( 1 + \left| \bm{\delta_{out}} - \bm{\delta_{out}}^* \right| \right) \ \ \text{otherwise} \\
      \end{aligned}
   \right.
\label{equ:offset_loss}
\end{equation}
where $\bm{\delta_{in}}$ and $\bm{\delta_{out}}$ refer to predictions and $\bm{\delta_{in}}^*$ and $\bm{\delta_{out}}^*$ are ground-truth. Note that $L_{off}$ is averaged separately for inside and outside objects due to the different formulations.

\subsection{Visual Properties Regression}
\label{sec:property}
We elaborate on the regression of visual properties including the 2D bounding boxes, dimensions, orientations, and keypoints of objects in this section.

\noindent\textbf{2D Detection.}
Since we do not represent objects as their 2D centers, we follow FCOS~\cite{tian2019fcos} to regress the distances from the representative point $\bm{x_r} = (u_r, v_r)$, which refers to $\bm{x_b}$ for inside and $\bm{x_I}$ for outside objects, to four sides of the 2D bounding box. If we denote the left-top corner as $(u_1, v_1)$ and the right-bottom corner as $(u_2, v_2)$, the regression target is then:
\begin{equation}
    \begin{aligned}
    l^* &= u_r - u_1, ~~r^* = u_2 - u_r, \\
    t^* &= v_r - v_1,\ ~~b^* = v_2 - v_r.
\end{aligned}
\end{equation}

GIoU loss~\cite{giou} is adopted for 2D detection since it is robust to scale changes.

\noindent\textbf{Dimension Estimation.}
Considering the small variance of object sizes within each category, we regress the relative changes with respect to the statistical average instead of absolute values. For each class $c$, the average dimension of the training set is denoted as $(\overline{h}_c, \overline{w}_c, \overline{l}_c)$. Assume the regressed log-scale offsets of dimensions are $(\delta_h, \delta_w, \delta_l)$ and the ground-truth dimensions are $(h^*, w^*, l^*)$, the L1 loss for dimension regression is defined as:
\begin{equation}
\begin{aligned}
   L_{dim} = \sum_{k \in \{h, w, l\}} \left|\overline{k}_{c} e^{\delta_{k}} - k^* \right|
\end{aligned}
\label{equ:dim}
\end{equation}

\begin{figure}[t]
\centering
\includegraphics[width=0.9\linewidth]{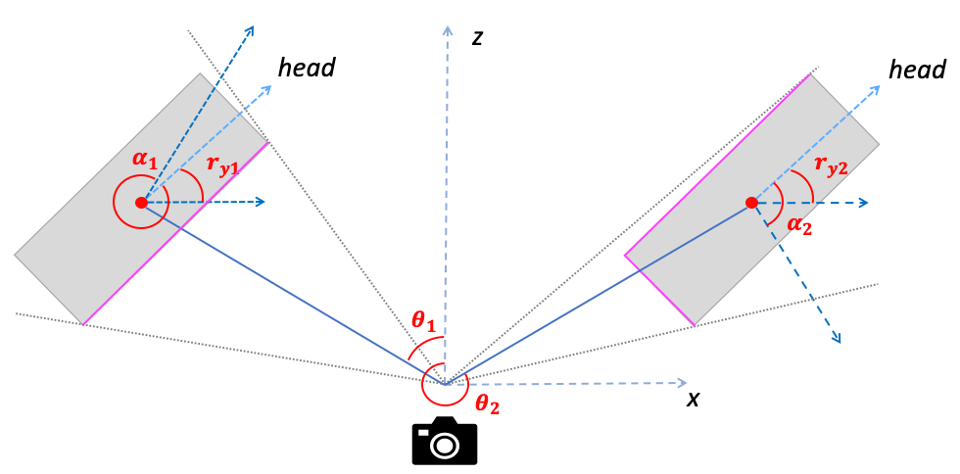}
\caption{$r_y$, $\alpha$, and $\theta$ are the global orientation, local orientation, and the viewing angle.}
\label{fig:orientation}
\end{figure}

\noindent\textbf{Orientation Estimation.}
The orientation can be represented as either the global orientation in the camera coordinate system or the local orientation relative to the viewing direction. For an object located at $(x, y, z)$, its global orientation $r_y$ and local orientation $\alpha$ satisfy~\eqref{equ:orien}:
\begin{equation}
    r_y = \alpha + \arctan(x / z)
    \label{equ:orien}
\end{equation}

As shown in Figure~\ref{fig:orientation}, objects with the same global orientations but different viewing angles will have different local orientations and visual appearances. Thus, we choose to estimate the local orientation with \textit{MultiBin} loss~\cite{chabot2017deep}, which divides the orientation range into $N_o$ overlapping bins so that the network can determine which bin an object lies inside and estimate the residual rotation w.r.t the bin center.

\begin{figure}[t]
\centering
\includegraphics[width=0.95\linewidth]{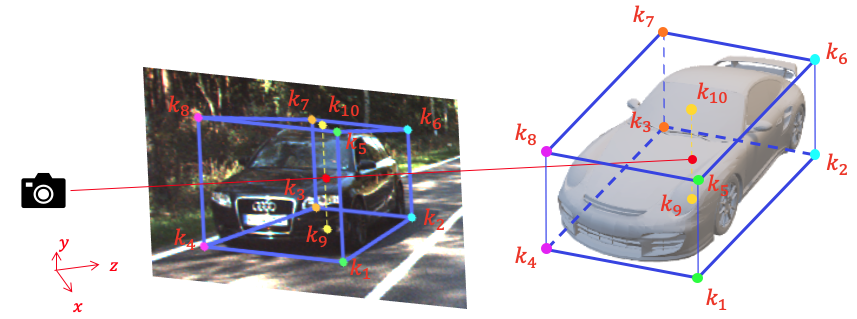}
\caption{Keypoints include the projections of eight vertexes, top center and bottom center of the 3D bounding box.}
\label{fig:keypoints}
\end{figure}

\noindent\textbf{Keypoint Estimation.}
As shown in Figure~\ref{fig:keypoints}, we define $N_k = 10$ keypoints for each object which include the projections of eight vertexes $\{\bm{k}_i, i=1...8\}$, bottom center $\bm{k}_9$ and top center $\bm{k}_{10}$ of the 3D bounding box. We regress the local offsets $\{\bm{\delta}_{ki} = \bm{k}_i - \bm{x_r}, i=1...N_k\}$ from $\bm{x_r}$ to $N_k$ keypoints with L1 loss:
\begin{equation}
   L_{key} = \frac{\sum_{i=1}^{N_k} I_{in}(\bm{k}_i) \left|\bm{\delta}_{ki} - \bm{\delta}^*_{ki} \right|}{\sum_{i=1}^{N_k} I_{in}(\bm{k}_i)}
   \label{equ:key_loss}
\end{equation}
where $\bm{\delta}^*_{ki}$ is the ground-truth and $I_{in}({\bm{k}_i})$ indicates whether the keypoint $\bm{k}_i$ is inside the image.

\subsection{Adaptive Depth Ensemble.}
\label{sec:adaptive_depth}
We formulate the estimation of object depth as an adaptive ensemble of $M + 1$ independent estimators, including direct regression and $M$ geometric solutions from keypoints. We first introduce these depth estimators and then present how we combine them with uncertainties.

\noindent\textbf{Direct Regression.}
To directly regress the object depth, we follow~\cite{chen2020cvpr, zhou2019objects} to transform the unlimited network output $z_o$ into the absolute depth $z_r$ with the inverse sigmoid transformation:
\begin{equation}
   z_r = \frac{1}{\sigma(z_o)}  - 1, \quad \sigma(x) = \frac{1}{1 + e^{-x}}
\end{equation}

To jointly model the uncertainty of the regressed depth, we follow~\cite{choi2019gaussian, kendall2017uncertainties, kendall2018multi} to utilize a modified L1 loss for depth regression:
\begin{equation}
   L_{dep} = \frac{\left| z_{r} - z^* \right|}{\sigma_{dep}} + \log(\sigma_{dep})
   \label{equ:l_d}
\end{equation}
where $\sigma_{dep}$ is the uncertainty of the regressed depth. When the model lacks confidence in its prediction, it will output a larger $\sigma_{dep}$ so that $L_{dep}$ can be reduced. The term $\log(\sigma_{dep})$ can avoid trivial solutions and encourage the model to be optimistic about accurate predictions.

\begin{figure}[t]
\centering
\includegraphics[width=0.9\linewidth]{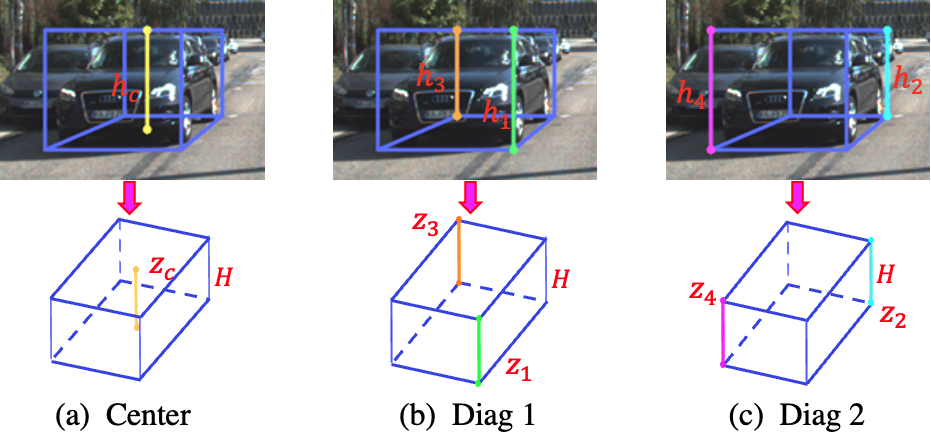}
\caption{The depth of a supporting line of the 3D bounding box can be computed with the object height and the line's pixel height. We split ten keypoints into three groups, each of which can produce the center depth independently.}
\label{fig:keypoint_to_depth}
\end{figure}

\noindent\textbf{Depth From Keypoints.}
\label{sec:depth_from_keypoints}
With known camera matrices, we can utilize the relative proportion between pixel height and estimated object height to compute the object depth, which is similar to~\cite{cai2020monocular}.
From our baseline model, the relative errors of predicted dimensions are 5.2\%, 6.1\%, and 11.8\% for height, width, and length. Therefore, solving depth from height is not only independent of orientation estimation but suffers less from the error of dimension estimation.
As shown in Figure~\ref{fig:keypoint_to_depth}, the estimated ten keypoints constitute five vertical supporting lines of the 3D bounding box. The depth $z_l$ of each vertical line can be computed from its pixel height $h_l$ and the object height $H$ as~\eqref{equ:h_to_depth}:
\begin{equation}
   z_l = \frac{f \times H}{h_l}
   \label{equ:h_to_depth}
\end{equation}
where $f$ is the camera's focal length. The depth of the center vertical line $z_c$ is exactly the object depth while averaging the depths of two diagonal vertical edges, namely $z_1$ and $z_3$ or $z_2$ and $z_4$, can also get the object depth. Therefore, the estimated ten keypoints are divided into three groups and generate respectively independent depths denoted as the center depth $z_c$, the $\text{diag}_1$ depth $z_{d_1}$ and the $\text{diag}_2$ depth $z_{d_2}$.

To further supervise the computed depths from keypoints and also model their uncertainties, we adopt the L1 loss with uncertainty as follows:
\begin{equation}
   L_{kd} = \sum_{k \in \{c, d_1, d_2\}} \left[\frac{\left| z_k - z^* \right|}{\sigma_k} + I_{in}(z_k) \log(\sigma_k)  \right]
\end{equation}
where $z^*$ is ground-truth and $I_{in}(z_k)$ indicates whether all keypoints used for computing $z_k$ are inside the image. Removing the log uncertainty for ``invalid" depths computed from invisible keypoints allows the model to be fully pessimistic so that these depths are down-weighted in the ensemble. Note that we also restrict the gradients from these invalid depths to only update the uncertainty.

\noindent\textbf{Uncertainty Guided Ensemble.}
Now that we have $M+1$ predicted depths $\{ z_i, i=1...M+1 \}$ and their uncertainties $\{ \sigma_i, i=1...M+1 \}$ from $M+1$ independent estimators, we propose to compute the uncertainty-weighted average, namely soft ensemble, as expressed in~\eqref{equ:soft}:
\begin{equation}
   z_{soft} = \left( \sum_{i=1}^{M+1} \frac{z_i}{\sigma_i} \right) / \left( \sum_{i=1}^{M+1} \frac{1}{\sigma_i} \right)
   \label{equ:soft}
\end{equation}

The soft ensemble can assign more weights to those more confident estimators while being robust to potential inaccurate uncertainties. We also consider the hard ensemble where the estimator with minimal uncertainty is chosen as the final depth estimation. The performances of two ensemble ways are compared in Section~\ref{sec:depth_combine}.

\noindent\textbf{Integral Corner Loss.}
As discussed in~\cite{qin2019monogrnet, simonelli2019disentangling}, the separate optimization of multiple subtasks cannot ensure the optimal cooperation among different components. Therefore, we also supervise the coordinates of eight corners $\{\bm{v}_i=(x_i, y_i, z_i), i=1,...,8\}$ from the predicted 3D bounding box, which is formed by the estimated dimension, orientation, offset, and soft depth $z_{soft}$, with L1 loss:
\begin{equation}
   L_{corner} = \sum_{i=1}^8 \left| \bm{v}_i - \bm{v}_i^* \right|
   \label{equ:corner_loss}
\end{equation}



\begin{table*}[t]
\small
\centering
   \begin{tabular}{| c  | c  | c | c | c | c | c | c | c | c | c | c | }
   \hline
   \multirow{2}{*}{Methods} & \multirow{2}{*}{Extra} & Time & \multicolumn{3}{c|}{Val, $\text{AP}_{3D}|_{R_{11}}$} &
   \multicolumn{3}{c|}{Val, $\text{AP}_{3D}|_{R_{40}}$} &
   \multicolumn{3}{c|}{Test, $\text{AP}_{3D}|_{R_{40}}$} \\
   & & (ms) & Easy & Mod & Hard & Easy & Mod & Hard & Easy & Mod & Hard\\
   \hline
   MonoPSR\cite{ku2019monopsr} & depth, LiDAR & 120   & 12.75 & 11.48 & 8.59  & -     & -     & -     & 10.76 & 7.25  & 5.85 \\
   UR3D\cite{UR3D} & depth & 120   & 28.05 & 18.76 & 16.55 & 23.24 & 13.35 & 10.15 & 15.58 & 8.61  & 6.00 \\
   AM3D\cite{Ma_2019_ICCV}  & depth & -     & 32.23 & 21.09 & 17.26 & 28.31 & 15.76 & 12.24 & 16.50 & 10.74 & 9.52 \\
   PatchNet\cite{PatchNet} & depth & -     & 35.10 & 22.00 & 19.60 & 31.60 & 16.80 & 13.80 & 15.68 & 11.12 & 10.17 \\
   DA-3Ddet\cite{DA-3Ddet} & depth, LiDAR & -     & 33.40 & 24.00 & 19.90 & -     & -     & -     & 16.80 & 11.50 & 8.90 \\
   D4LCN\cite{ding2020learning} & depth & -     & 26.97 & 21.71 & 18.22 & 22.32 & 16.20 & 12.30 & 16.65 & 11.72 & 9.51 \\
   Kinem3D\cite{brazil2020kinematic} & multi-frames & 120   & -     & -     & -     & 19.76 & 14.10 & 10.47 & 19.07 & 12.72 & 9.17 \\
   \hline
   FQNet\cite{liu2019deep} & -     & -     & 5.98  & 5.50  & 4.75  & -  & - & -  & 2.77  & 1.51  & 1.01 \\
   MonoGRNet\cite{qin2019monogrnet} & -     & 60    & 13.88 & 10.19 & 7.62  & -     & -     & -     & 9.61  & 5.74  & 4.25 \\
   MonoDIS\cite{simonelli2019disentangling} & -     & 100   & 18.05 & 14.98 & 13.42 & -     & -     & -     & 10.37 & 7.94  & 6.40 \\
   M3D-RPN\cite{Brazil_2019_ICCV} & -     & 160   & 20.27 & 17.06 & 15.21 & 14.53 & 11.07 & 8.65  & 14.76 & 9.71  & 7.42 \\
   SMOKE\cite{liu2020smoke} & -  & \textbf{30} & 14.76 & 12.85 & 11.50 & -     & -     & -     & 14.03 & 9.76  & 7.84 \\
   MonoPair\cite{chen2020cvpr} & -     & 57    & -     & -     & -     & 16.28 & 12.30 & 10.42 & 13.04 & 9.99  & 8.65 \\
   RTM3D\cite{li2020rtm3d} & -     & 55    & 20.77 & 16.86 & 16.63 & -     & -     & -     & 14.41 & 10.34 & 8.77 \\
   Movi3D\cite{movi3D}  & -     & 45    & -     & -     & -     & 14.28 & 11.13 & 9.68  & 15.19 & 10.90 & 9.26 \\
   \hline
   Ours & - & 35 & \textbf{28.17} & \textbf{21.92} & \textbf{19.07} & \textbf{23.64} & \textbf{17.51} & \textbf{14.83} & \textbf{19.94} & \textbf{13.89} & \textbf{12.07} \\
   \hline
\end{tabular}
\caption{Quantitative results for Car on KITTI \textit{val}/\textit{test} sets, evaluated by $AP_{3D}$. ``Extra" lists the required extra information for each method. We divide existing methods into two groups considering whether they utilize extra information and sort them according to their performance on the moderate level of the \textit{test} set within each group.}
\label{tab:3d}
\end{table*}

\begin{table}[t]
\footnotesize
\centering
\begin{tabular}{|c|c|c|c|c|c|c|}
\hline
\multirow{3}{*}{Methods} & \multicolumn{6}{c|}{Test, $\text{AP}_{3D}|_{R_{40}}$}\\
\cline{2-7}
& \multicolumn{3}{c|}{Pedestrian} & \multicolumn{3}{c|}{Cyclist}\\
\cline{2-7}
& Easy & Mod & Hard & Easy & Mod & Hard\\
\hline
M3D-RPN\cite{Ma_2019_ICCV} & 4.92 & 3.48 & 2.94 & 0.94 & 0.65 & 0.47\\
Movi3D\cite{movi3D} & 8.99 & 5.44 & 4.57 & 1.08 & 0.63 & 0.70\\
MonoPair\cite{chen2020cvpr} & \textbf{10.02} & \textbf{6.68} & \textbf{5.53} & 3.79 & 2.12 & 1.83\\
\hline
Ours & 9.43 & 6.31 & 5.26 & \textbf{4.17} & \textbf{2.35} & \textbf{2.04}\\
\hline
\end{tabular}
\caption{Quantitative results for Pedestrian and Cyclist on KITTI \textit{test} set.}
\label{tab:other_class}
\end{table}

\section{Experiments}
The proposed method is evaluated on KITTI 3D Object Detection benchmark~\cite{geiger2012we}, which includes 7481 images for training and 7518 images for testing. We follow~\cite{chen20153d} to split the training images into \textit{train} (3712) and \textit{val} (3769) sets. Detection results are evaluated on three levels of difficulty: easy, moderate, and hard, which are defined by the bounding box height, occlusion, and truncation. All our reported results are produced by models that jointly detect multi-classes, including Car, Pedestrian, and Cyclist. Note that results for KITTI Bird's Eye View benchmark will be provided in the supplementary material for reference.

\subsection{Implementation Details}
We adopt the same modified DLA-34~\cite{dla} as our backbone network following~\cite{chen2020cvpr, liu2020smoke, zhou2019objects}. All input images are padded to the same size of $384\times1280$. Every prediction head attached to the backbone consists of one $3 \times 3 \times 256$ conv layer, BatchNorm~\cite{ioffe2015batch}, ReLU, and another $1\times 1 \times c_o$ conv layer, where $c_o$ is the output size. The edge fusion module has similar settings except using 1D conv layer and empirically removing the ReLU activation. For \textit{MultiBin} loss~\cite{deep3Dbox}, we use four bins centered at $[0, \frac{\pi}{2}, \pi, -\frac{\pi}{2}]$. The model is trained using AdamW~\cite{adamw} optimizer with the initial learning rate as 3e-4 and weight decay as 1e-5. We train the model for 34k iterations with a batchsize of 7 on a single RTX 2080Ti GPU and the learning rate is divided by 10 at 22k and 30k iterations. The random horizontal flip is adopted as the only data augmentation.

\subsection{Evaluation Metrics}

The detection is evaluated by the average precision of 3D bounding boxes $\text{AP}_{3D}$. For the \textit{val} set, we report both $\text{AP}_{3D}|_{R_{11}}$ and $\text{AP}_{3D}|_{R_{40}}$ for a comprehensive comparison with previous studies. For the \textit{test} set, the $\text{AP}_{3D}|_{R_{40}}$ results from the test server are reported.
The IoU thresholds for $\text{AP}_{3D}$ are 0.7 for Car and 0.5 for Pedestrian and Cyclist.

\subsection{Quantitative Results}

In Table~\ref{tab:3d}, we conduct a comprehensive comparison between our proposed method and existing arts on the \textit{val} and \textit{test} sets of KITTI benchmark for Car. Without bells and whistles, our method outperforms all prior methods that only take monocular images as input. For $\text{AP}_{3D}|_{R_{40}}$ on the \textit{val} set, our method is 45\%, 42\% and 42\% higher than the second-best method MonoPair~\cite{chen2020cvpr} on three levels of difficulty.
For the \textit{test} set, our proposed method surpasses all existing methods, including those with extra information. The significant improvement on hard samples demonstrates that our method can effectively detect those heavily truncated objects, which are crucial for practical applications.
We further show the results of Pedestrian and Cyclist on the \textit{test} set in Table~\ref{tab:other_class}. Our method outperforms M3D-RPN~\cite{Brazil_2019_ICCV} and Movi3D~\cite{movi3D} while achieving comparable performance with MonoPair~\cite{chen2020cvpr}.
Finally, our method is also much faster than most existing methods, allowing for real-time inference. To sum up, our proposed framework achieves a state-of-the-art trade-off between performance and latency.

\subsection{Ablation Study}

\subsubsection{Decoupled Representations} \label{sec:repre_ablation}

In Table~\ref{tab:ablation_center}, we compare various representations for inside and outside objects and validate the improvement from separate offset losses, namely the decoupled loss, and the edge fusion module. The second row which represents all objects with $\bm{x_b}$ is considered as the baseline. All models directly regress the object depth without ensemble.

We observe that:
(1) Simply discarding outside objects can improve the performance compared to the baseline, demonstrating the necessity of decoupling outside objects.
(2) Identifying inside objects as their projected 3D centers $\bm{x_c}$ is better than the 2D centers $\bm{x_b}$, possibly because the offsets from $\bm{x_b}$ to $\bm{x_c}$ are irregular and hard to learn.
(3) The decoupled optimization of inside and outside offsets and the edge fusion module are crucial for the remarkable improvement on moderate and hard samples, where the heavily truncated objects belong.
(4) Compared with $\bm{x_{cc}}$ and $\bm{x_{cb}}$ derived by clamping $\bm{x_c}$ and $\bm{x_b}$ to the image edge as shown in Figure~\ref{fig:center_fig}(a), the proposed intersection $\bm{x_I}$ is a more effective representation for outside objects.

\begin{table}[t]
\centering
   \footnotesize
   \begin{tabular}{|c|c|c|c|c|c|c|}
   \hline
   \multicolumn{2}{|c|}{Representation} & Decoupled & Edge & \multicolumn{3}{c|}{Val, $\text{AP}_{\text{3D}}|_{R_{40}}$}\\
   \cline{1-2}\cline{5-7}
   Inside & Outside & Loss & Fusion & E & M & H \\
   \hline
   $\bm{x_b}$ & - & \multicolumn{2}{c|}{\multirow{2}{*}{}} & 13.5 & 10.5 & 8.9 \\
   $\bm{x_b}$ & $\bm{x_b}$  & \multicolumn{2}{c|}{} & 13.0 & 9.4 & 7.6 \\
   \hline
   $\bm{x_c}$ & - & & & 15.3 & 11.0 & 9.6  \\
   $\bm{x_c}$ & $\bm{x_I}$ & & & 13.9 & 10.2 & 8.9 \\
   $\bm{x_c}$ & $\bm{x_I}$ & $\surd$ & & 14.2 & 11.7 & 9.8\\
   $\bm{x_c}$ & $\bm{x_I}$ & & $\surd$ & 14.6 & 11.7 & 9.7\\
   $\bm{x_c}$ & $\bm{x_I}$ & $\surd$ & $\surd$ & 15.9 & \textbf{12.6} & \textbf{11.4} \\
   \hline
   $\bm{x_c}$ & $\bm{x_{cc}}$ & $\surd$ & $\surd$ & 12.1 & 9.8 & 8.6\\
   $\bm{x_c}$ & $\bm{x_{cb}}$ & $\surd$ & $\surd$ & \textbf{16.2} & 12.0 & 10.2\\
   \hline
   \end{tabular}
\caption{Ablation study on decoupled representations.}
\label{tab:ablation_center}
\end{table}

\begin{figure*}
\centering
\includegraphics[width=0.95\linewidth]{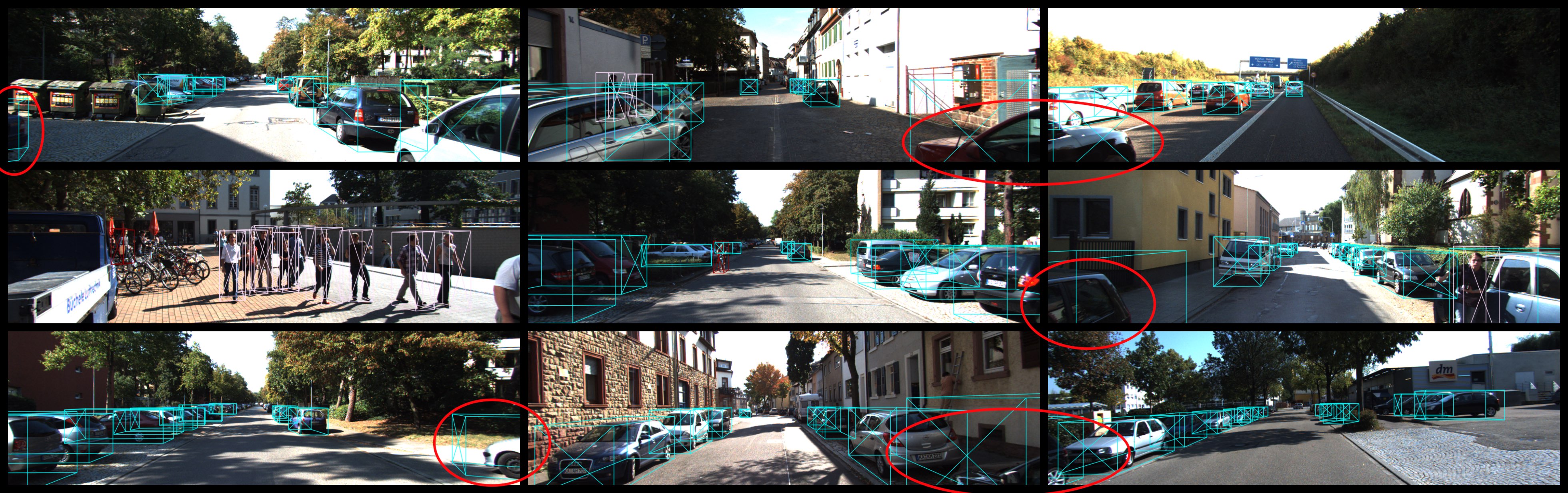}
\caption{\textbf{Qualitative Results.} We visualize the results of 3D object detection on KITTI \textit{val} set, where predicted cars, pedestrians, and cyclists are represented in cyan, light pink, and red boxes. We use red ovals to emphasize those heavily truncated objects.}
\label{fig:results}
\end{figure*}

\subsubsection{Object Depth Estimation}
\label{sec:depth_ablation} 
We compare different methods for object depth estimation in Table~\ref{tab:ablation_depth}. ``Direct Regression" refers to the best model in Table~\ref{tab:ablation_center} with our decoupled representations but without estimating keypoints. ``Keypoints" replaces the depth branch with keypoint prediction and solves the object depth from geometry as in Section~\ref{sec:depth_from_keypoints}.
The regressed depth performs slightly better than the keypoints-based solutions. The import of uncertainties significantly improves both methods because it allows the model to neglect difficult outliers and focus on most moderate objects. By contrast, our adaptive depth ensemble method simultaneously performs both predictions and further combines them with the uncertainty-guided weights, outperforming all individual methods by an obvious margin.

\begin{table}
\footnotesize
\centering
   \begin{tabular}{|c|c|c|c|}
   \hline
   \multirow{2}{*}{Depth Method} & \multicolumn{3}{c|}{Val, $\text{AP}_{\text{3D}}|_{R_{40}}$}\\
   \cline{2-4}
    & Easy & Mod & Hard\\
   \hline
   Direct Regression & 15.86 & 12.60 & 11.38 \\
   Direct Regression + $\sigma$ & 19.63 & 14.83 & 13.25 \\
   Keypoints & 15.45 & 12.18 & 10.73 \\
   Keypoints + $\sigma$ & 18.42 & 14.76 & 12.49\\
   Adaptive Ensemble & \textbf{23.64} & \textbf{17.51} & \textbf{14.83} \\
   \hline
   \end{tabular}
\caption{Ablation study on object depth estimation.}
\label{tab:ablation_depth}
\end{table}

\subsection{Depth Combination}
\label{sec:depth_combine}
To further understand the effectiveness of the proposed depth ensemble, we compare the performance of each estimator and the combined depth from the ensemble model in Table~\ref{tab:ense}. It can be observed that the joint learning consistently improves the performance of each depth estimator compared with results in Table~\ref{tab:ablation_depth}, which can owe to enhanced feature learning. The combined depth from soft ensemble outperforms every individual estimator, especially for the moderate level of Car and all levels of Pedestrian. The hard ensemble is inferior, possibly due to its sensitivity to the mismatch between the actual depth error and the estimated uncertainty.
We also provide the performance from the oracle depth which means the most accurate estimator is always chosen for each object by an oracle. It can be considered as the ideal upper bound of our depth ensemble. To match a predicted object with a ground-truth object, we require their 2D IoU to be larger than 0.5. We notice that our soft ensemble is very close to the oracle performance on Pedestrian, demonstrating the effectiveness of our proposed combination method. On the other hand, the oracle performance for Car reveals the enormous potential of combining different depth estimators, which can be left for future work.

\begin{table}
\centering
   \footnotesize
   \begin{tabular}{|c|c|c|c|c|c|c|}
   \hline
   \multirow{3}{*}{Estimator} & \multicolumn{6}{c|}{Val, $\text{AP}|_{R_{40}}$}\\
   \cline{2-7}
   & \multicolumn{3}{c|}{Car, IoU$>$0.7} & \multicolumn{3}{c|}{Pedestrian, IoU$>$0.5}\\
   \cline{2-7}
   & Easy & Mod & Hard & Easy & Mod & Hard\\
   \hline
   Regression & \underline{23.41} & 16.83 & 14.59 & 7.39 & 5.81 & 4.54\\
   \textit{Key}: Center & 23.29 & \underline{16.84} & \underline{14.72} & 7.40 & 5.74 & 4.54\\
   \textit{Key}: $\text{Diag}_1$ & 23.13 & 16.70 & 14.50 & 7.30 & 5.64 & 4.52\\
   \textit{Key}: $\text{Diag}_2$ & 23.35 & 16.81 & 14.63 & 7.38 & 5.76 & 4.56\\
   Hard Ensemble & 22.58 & 16.80 & 14.58 & \underline{7.51} & \underline{6.38} & \underline{4.64}\\
   Soft Ensemble & \textbf{23.64} & \textbf{17.51} & \textbf{14.83} & \textbf{8.16} & \textbf{6.45} & \textbf{5.16}\\
   \hline
   Oracle & 26.28 & 19.98 & 17.07 & 8.54 & 6.72 & 5.55\\
   \hline
   \end{tabular}
\caption{Quantitative analysis for the adaptive ensemble of depth estimators.}
\label{tab:ense}
\end{table}



\subsection{Qualitative Results}
From the qualitative results shown in Figure~\ref{fig:results}, our proposed framework can produce superior performance for ordinary objects in various street scenes. As highlighted by the red ovals, we can also successfully detect some extremely truncated objects which are crucial for the safety of autonomous driving, demonstrating the effectiveness of decoupling truncated objects.

\section{Conclusion}
In this paper, we have proposed a novel framework for monocular 3D object detection which flexibly handles different objects. We observe the long-tail distribution of truncated objects and explicitly decouple them with the proposed edge heatmap and edge fusion module.
We also formulate the object depth estimation as an uncertainty-guided ensemble of multiple approaches, leading to more robust and accurate predictions.
Experiments on KITTI benchmark show that our method significantly outperforms all existing competitors. Our work sheds light on the importance of flexibly processing different objects, especially for the challenging monocular 3D object detection.

\section*{Acknowledgement}
This work was supported in part by the National Natural Science Foundation of China under Grant U1713214, Grant U1813218, Grant 61822603, in part by Beijing Academy of Artificial Intelligence (BAAI), and in part by a grant from the Institute for Guo Qiang, Tsinghua University.

\newpage

{\small
\bibliographystyle{ieee_fullname}
\bibliography{egbib}
}






\end{document}